\documentclass[letterpaper]{article} 
\usepackage{aaai25}  
\usepackage{times}  
\usepackage{helvet}  
\usepackage{courier}  
\usepackage[hyphens]{url}  
\usepackage{graphicx} 
\usepackage{natbib}  
\usepackage{caption} 
\usepackage{refstyle}
\usepackage{algorithm}
\usepackage{algorithmic}
\usepackage{amsmath} 
\usepackage{graphicx}
\usepackage{amssymb}
\usepackage{booktabs,subcaption,amsfonts,dcolumn}
\usepackage{bbold}
\usepackage{xcolor}
\usepackage{caption}
\usepackage{multirow}
\usepackage{comment}
\usepackage{newfloat}
\usepackage{listings}
\DeclareCaptionStyle{ruled}{labelfont=normalfont,labelsep=colon,strut=off} 
\floatstyle{ruled}
\newfloat{listing}{tb}{lst}{}
\floatname{listing}{Listing}
\title{Test-time Conditional Text-to-Image Synthesis Using Diffusion Models}
\author{
    Tripti Shukla,
    Srikrishna Karanam,
    Balaji Vasan Srinivasan
}
\affiliations{
    Adobe Research, Bengaluru India\\

    \{trshukla, skaranam, balsrini\}@adobe.com
}

\usepackage{bibentry}
\usepackage{cleveref}

\begin{document}


\twocolumn[{
\renewcommand\twocolumn[1][]{#1}%
\maketitle
\begin{center}
 \centering
 \captionsetup{type=figure}
 \vspace{-6mm}
\includegraphics[width=\textwidth]
 {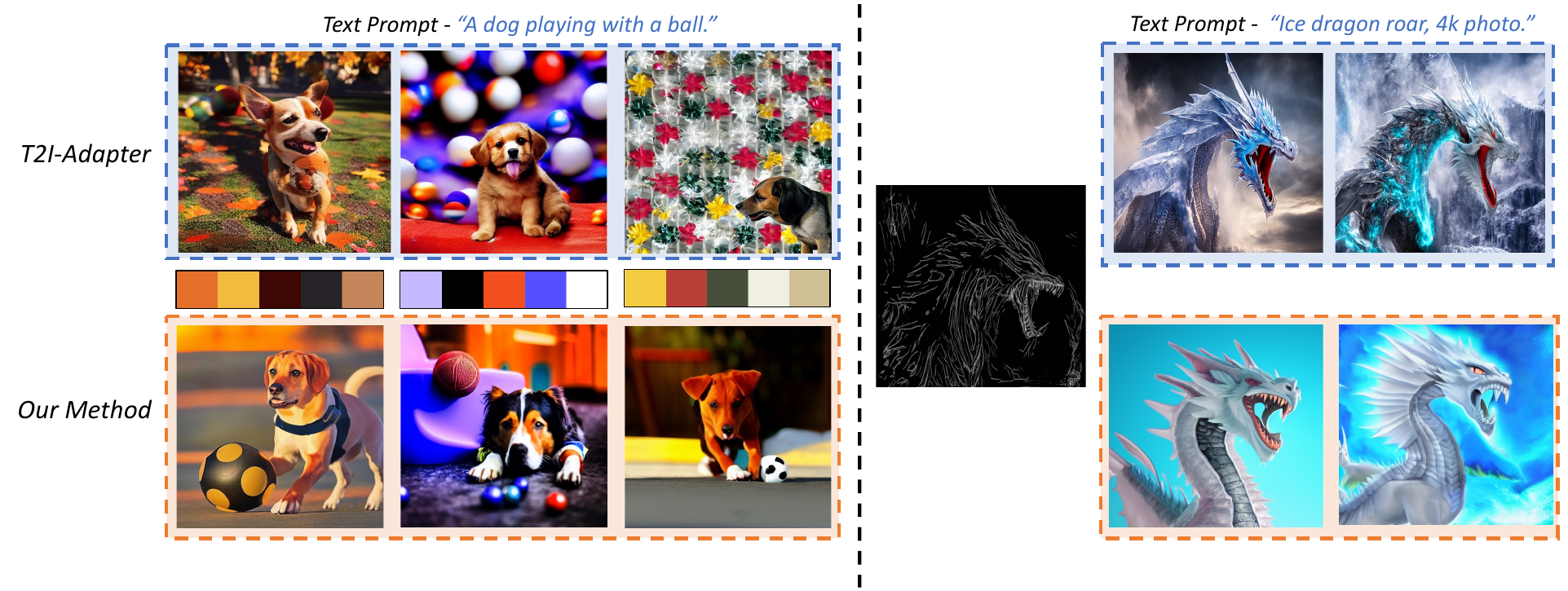}
 \caption{We propose \textbf{TINTIN}, a training-free method to generate images with an extra condition using text-to-image diffusion models preserving the generation quality and diversity in the images. Our method performs well in comparison with training-dependent models like T2I-Adapter \cite{mou2023t2iadapter}. The first column illustrates the results of color conditioning, while the second column showcases the outcomes of edge conditioning. }
 \label{fig:teaser_qual}
\end{center}
}]

\begin{abstract}
We consider the problem of conditional text-to-image synthesis with diffusion models. Most recent works need to either finetune specific parts of the base diffusion model or introduce new trainable parameters, leading to deployment inflexibility due to the need for training. To address this gap in the current literature, we propose our method called \textit{\textbf{TINTIN}: \textbf{T}est-t\textbf{i}me Co\textbf{n}ditional \textbf{T}ext-to-\textbf{I}mage Synthesis using Diffusio\textbf{n} Models} which is a new training-free test-time only algorithm to condition text-to-image diffusion model outputs on conditioning factors such as color palettes and edge maps. In particular, we propose to interpret noise predictions during denoising as gradients of an energy-based model, leading to a flexible approach to manipulate the noise by matching predictions inferred from them to the ground truth conditioning input. This results in, to the best of our knowledge, the first approach to control model outputs with input color palettes, which we realize using a novel color distribution matching loss. We also show this test-time noise manipulation can be easily extensible to other types of conditioning, e.g., edge maps. We conduct extensive experiments using a variety of text prompts, color palettes, and edge maps and demonstrate significant improvement over the current state-of-the-art, both qualitatively and quantitatively.
\end{abstract}

\section{Introduction} 
\label{section:intro}

With the dramatic rise in the capabilities of text-to-image diffusion models \cite{rombach2022highresolution,saharia2022photorealistic} in generating creative imagery from text prompts, much recent effort has been expended in controlling the outputs generated by these models. For instance, methods like ControlNet \cite{liu2023control} presented ways to condition the outputs of text-to-image diffusion models on a wide variety of conditioning factors such as color, edge, depth, etc. Since then, many other methods have been proposed \cite{mou2023t2iadapter} that seek to train additional modules to incorporate the extra conditioning input along with the text prompt. A recent contribution in the realm of training-free approaches is presented in the work by Yu et al. (referred as FreeDoM from here onwards)\cite{yu2023freedom}. This approach empowers conditional generation without the need for training or fine-tuning additional models. It excels in handling relatively simple conditions such as poses, segmentation maps, and similar factors.

\begin{figure}[h]
\centering
\includegraphics[width = \linewidth]{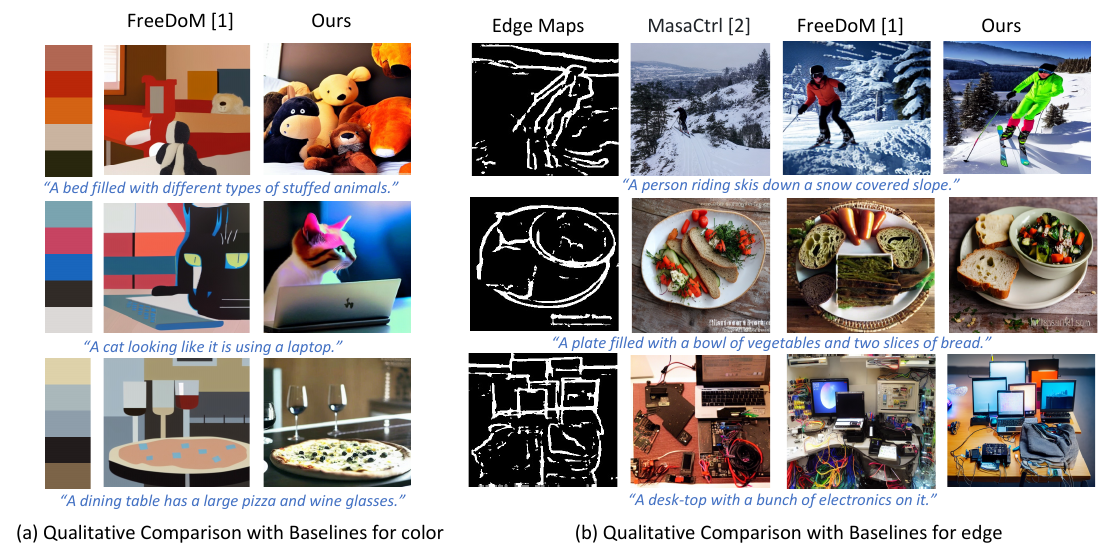}
\caption{Comparison of TINTIN with FreeDoM \cite{yu2023freedom} for color and edge conditions.}
\vspace{-5mm}
\label{fig:compare_free}
\end{figure}

While the aforementioned techniques have shown excellent conditioning capabilities, there are some limitations. First, these methods introduce extra trainable parameters/modules that necessarily require an extra training step before denoising at test time, leading to inflexibility for practical deployment since each time there is a new conditioning factor, one needs to retrain the model. Next, since these methods have extra training modules (and these are generally small-scale neural networks), they would work well in the distribution of the data that was used to train them. If we want the conditioning to work in a wide variety of scenarios not limited to the training distribution (which is a very practically relevant scenario), that will likely mean gathering more data to finetune these extra modules, leading to more deployment inflexibility. The training-free method proposed by FreeDoM\cite{yu2023freedom} presents a generic strategy that works well for face-related data, as demonstrated in the paper, but it encounters challenges when applied to a larger data domain with diverse images. While it effectively handles simpler conditions such as poses and segmentation maps, it struggles with more intricate controls like color palettes and edge maps. These limitations primarily arise from the absence of specific loss functions and the precise application of conditioning at optimal time intervals, along with other algorithmic adjustments tailored to fine-grained structural features within extensive data domains. For example, colorization details are determined in the intermediate part of the sampling interval \cite{agarwal2023imageworthmultiplewords}, where the use of a color-specific loss function can ensure that these details are accurately captured during generation. Additionally, its generalizability remains largely confined to human faces, underscoring the need for advancements to accommodate a broader range of conditions for diverse set of data. Consequently, the key question we ask in this paper is- \textit{can we devise a test-time only adaptation technique that can achieve the desired conditionings without having to retrain/update base model parameters for large data domains?}

To address the aforementioned question and the gaps in the current literature, we propose methods to manipulate the intermediate representations produced by the diffusion model during the denoising process at test time. We propose new loss functions for the specific tasks that help guide the intermediate noise predictions towards the desired state which when decoded give us the expected conditioned model outputs. We present two specific instantiations of our approach with color conditioning in the form of color palettes and edge conditioning in the form of edge maps. Comparing with FreeDoM \cite{yu2023freedom}, we propose a novel color loss function that promotes alignment between the target color distribution and the generated image's color distribution. We also propose a iterative sampling strategy specific to color and edge conditioning during test-time image generation.
Figure \ref{fig:compare_free} clearly establishes that the above contributions are crucial in the success of our approach in generating color and edge conditioned images as opposed to FreeDoM\cite{yu2023freedom}. To the best of our knowledge, this is the first approach that conditions model outputs based on input color palettes without any retraining at test time. Figure \ref{fig:teaser_qual} illustrates the generations from our training-free method against a training-dependent model called T2I-Adapter \cite{mou2023t2iadapter}. While we show results with color and edge inputs, given our approach only needs test-time adaptation, it can easily be extended to other kinds of conditioning given suitable ways to compute the predictions necessary to compute the resulting loss.



To summarize, the key contributions of this paper include:
\begin{itemize}
    \item We present a test-time-only approach to condition diffusion model outputs with external conditioning factors like color palettes and edge maps by interpreting noise predictions as the gradients of an energy-based model. 
    \item To the best of our knowledge, we present the first approach to condition diffusion models with color palettes without any retraining and only using a test-time manipulation of the noise predictions by matching the color distributions of the input palette and the prediction at a particular denoising step.
    \item We introduce a novel loss function $L_{DS}$ to optimize the alignment between the generated color distribution and the target palette distribution, enhancing the fidelity of the synthesized images.
    \item We present an iterative sampling strategy optimized for color and edge conditioning for test-time conditional generation.
\end{itemize}

\vspace{-4mm}
\section{Related Work}
Text-to-Image generation aims at generating realistic images from input text prompts. To address this task, there has been significant amount of work in the past. Generative Adversarial Networks (GANs) \cite{huang2017stacked,radford2016unsupervised,zhao2017energybased,karras2018progressive} have shown impressive performance in image generation tasks but these models suffer from some key challenges like training instability and poor generalizability.

In the realm of text-to-image generation using diffusion models, several works have been dedicated to generating high-quality images based on given text prompts \cite{dhariwal2021diffusion, wang2004image, ho2020denoising, gal2022image}. A recent trend in the diffusion model domain focuses on conditional image generation \cite{meng2022sdedit, Avrahami_2022, ramesh2022hierarchical}. Some approaches \cite{Avrahami_2022, brooks2023instructpix2pix, gafni2022makeascene, kawar2023imagic, kim2022diffusionclip, hertz2022prompttoprompt} address text guidance by manipulating prompts or modifying cross-attention. Other methods, such as T2I-Adapter \cite{mou2023t2iadapter}, Control-Net \cite{liu2023control}, and UniControlNet \cite{zhao2023unicontrolnet}, introduce additional conditions like pose, color, or segmentation maps to guide image generation alongside text prompts. However, these methods commonly rely on training or fine-tuning additional models, limiting their generalizability and usability across diverse conditions, and incurring associated training costs that impede broad applicability.

\begin{figure}[h]
 \centering
 \captionsetup{type=figure}
 \includegraphics[width = \linewidth]
 {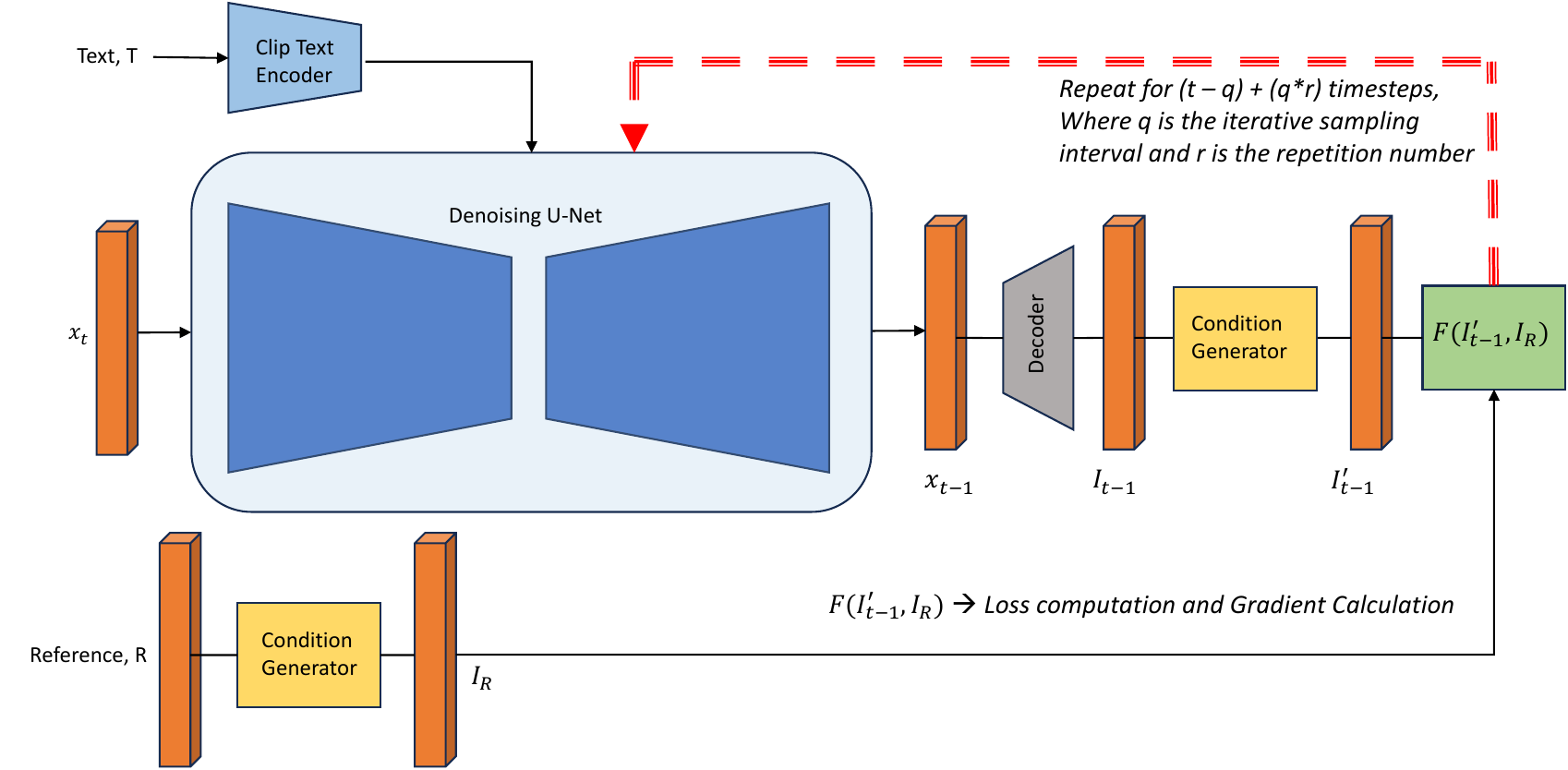}
 \caption{The overall architecture of our training-free approach, TINTIN. Reference, R is a list of hex values for color control and a reference image for edge control, $x_t$ is the input latent code generated from random Gaussian Distribution, $x_{t-1}$ is the latent code at timestep $t-1$ and $I_{t-1}$ is the denoised RGB image. \textit{Condition Generator network} projects the condition control and the denoised image in the same space denoted by $I_R$ and $I'_{t-1}$ respectively.}
 \label{fig:architecture}
 \vspace{-2mm}
\end{figure}

To address these challenges, some training-free approaches \cite{yu2023freedom,parmar2023zeroshot,bansal2023universal}, for conditional generation have been proposed. These methods eliminate the need for training or fine-tuning additional models, offering a more versatile and accessible solution. However, existing training-free approaches exhibit limitations when dealing with more nuanced conditions, such as color palettes and edge maps. Moreover, their performance struggles in large-data domains, and their generalizability is often confined to specific conditions like human faces.

\section{Our Proposed Approach}
\label{sec:proposed_approach}

As discussed in \ref{section:intro}, existing methods like \cite{mou2023t2iadapter,liu2023control} train models for conditional image generation. Recent works such as \cite{yu2023freedom} offer conditioning during inference, avoiding the need to train separate models. However, these methods only handle simple conditions like pose, style, and text, and struggle with large datasets like Imagenet.

For precise control over image generation, especially for detailed conditions like color and layout during test-time, conditional generation models are essential. Song et al. \cite{song2021scorebased} introduced a score-based model using Stochastic Differential Equation (SDE) for conditional generation. Their conditional reverse-time SDE, estimated from unconditional scores, significantly improves conditional synthesis. This inspires our exploration of score-based models for optimizing conditional generation.

We introduce TINTIN, which controls fine-grained conditions like color and edge maps and can generate images from large datasets. Our first contribution is \textit{Inference-time Conditional Generation}, generating images based on color palettes and edge maps during inference. For color conditioning, given a text prompt $t$ and a color palette $p$, our method generates diverse images based on $t$ and conditioned on $p$. For edge conditioning, given a text prompt $t$ and edge maps $e$, our method generates images matching the structure of $e$. Our second contribution proposes various loss functions for color and edge control. Our third contribution analyzes the iterative sampling strategy for color and edge conditioning.

\subsection{Preliminaries}
\textbf{Latent Diffusion Model.} Diffusion models \cite{sohldickstein2015deep,ho2020denoising} are probabilistic models for generating samples from a target data distribution. The generative process starts with a sample $x_0 \sim p(x_0)$ and gradually denoises it over $T$ iterations. This process transforms a normally distributed random variable $x_T \sim \mathcal{N}(0,I)$ into a series of intermediate samples $x_T, x_{T-1}, \ldots, x_0$, each with less noise. A neural network $\epsilon_{\theta}(x_t, t)$ determines the noise reduction at each step.

\textbf{Conditional Score Function.} \label{sec:cond_score} We overview score-based diffusion models (SBDMs) and conditional generation using them. SBDMs \cite{song2020generative,song2022solving} are a type of diffusion model based on score theory. They estimate the score function \(\nabla_{x_t}\log p(x_t)\) using a score estimator $s({x_t}, {t})$, where $x_t$ is noisy data and $t$ is the time-step. The score function provides the gradient of the log-likelihood of the data. During sampling, SBDMs derive $x_{t-1}$ from $x_t$ using this score function.

As discussed in Section \ref{sec:proposed_approach}, score-based models enable conditional generation since the conditional reverse-time SDE \cite{song2022solving} can be estimated from unconditional scores. The score function can be modified as \(\nabla_{x_t} \log p(x_t | c)\), incorporating a given condition $c$. Using Bayes' theorem \cite{bayes1763essay}, the conditional score function can be written as:
\begin{equation}
\nabla_{\mathbf{x}_t}\log p(\mathbf{x}_t|\mathbf{c}) = \nabla_{\mathbf{x}_t}\log p(\mathbf{x}_t) + \nabla_{\mathbf{x}_t}\log p(\mathbf{c}|\mathbf{x}_t) \label{eq:condition_score_theory}
\end{equation}
where \(\nabla_{x_t}\log p(x_t)\) is estimated from a pre-trained unconditional score estimator $\mathbf{s({x_t}, {t})}$ and \(\nabla_{x_t} \log p(c | x_t)\) enforces the condition in the model. The second term \(\nabla_{x_t} \log p(c | x_t)\) serves as a correction gradient, steering $x_t$ towards data that matches the condition $c$. Time-dependent classifiers \cite{liu2023control,zhao2022egsde,nichol2021glide,dhariwal2021diffusion} are trained to compute this correction gradient for precise conditional guidance.

\textbf{Energy Guided Diffusion.} We model the correction gradient \(\nabla_{x_t} \log p(c | x_t)\) as an energy function:
\begin{equation}
p(\mathbf{c}|\mathbf{x}_t) = \frac{\exp\{-\lambda \mathcal{E}(\mathbf{c}, \mathbf{x}_t)\}}{\int_{\mathbf{c \in C}}\exp\{-\lambda \mathcal{E}(\mathbf{c}, \mathbf{x}_t)\}} \label{eq:energy_function}
\end{equation}
where $\mathbf{c}$ is the condition, and $\mathbf{\lambda}$ is the positive temperature coefficient. The energy function $-E(c,\mathbf{x}_t)$ measures the alignment between $\mathbf{c}$ and $\mathbf{x_t}$, producing smaller values when they align better. Thus, the correction gradient is:
\begin{equation}
\nabla _{\mathbf {x}_t}\log p(\mathbf {c}|\mathbf {x}_t)\propto -\nabla _{\mathbf {x}_t}\mathcal {E}(\mathbf {c}, \mathbf {x}_t), \label{eq:score_approx_energy}
\end{equation}
called energy conditioning. Using the standard DDPM equation \cite{ho2020denoising}, Eq. \ref{eq:condition_score_theory}, and Eq. \ref{eq:score_approx_energy}, the conditional sampling formula is:
\begin{equation}
\mathbf{x}_{t-1} = \mathbf{r}_t - \alpha_t \nabla_{\mathbf{x}_t} \mathcal{E}(\mathbf{c}, \mathbf{x}_t), \label{eq:energy_samp} 
\end{equation}
where $\mathbf{r_t}$ is the standard DDPM sampling formula and $\mathbf{\alpha_t}$ is the learning rate of energy conditioning.

\textbf{Time Independent Energy Approximation.} Some methods \cite{yu2023freedom} approximate energy conditioning using time-independent distance functions due to many pretrained functions for clean data $\mathbf{x_0}$:
\begin{equation}
    \mathcal {D}_{\boldsymbol {\phi }}(\mathbf {c}, \mathbf {x}_t, t)\approx \mathbb {E}_{p(\mathbf {x}_0|\mathbf {x}_t)}[\mathcal {D}_{\boldsymbol {\theta }}(\mathbf {c}, \mathbf {x}_0)]. \label {eq:estimate_noisy_s_using_clean_x0t} 
\end{equation}
where $\mathcal {D}_{\boldsymbol {\phi }}(\mathbf {c}, \mathbf {x}_t, \mathbf{t})$ approximates energy conditioning with pretrained parameter $\mathbf{\phi}$. $\mathcal {D}_{\boldsymbol {\theta }}(\mathbf {c}, \mathbf {x}_0)$ denotes time-independent distance networks for clean data with pretrained parameter $\mathbf{\theta}$. The distance between $\mathbf{x_t}$ and $\mathbf{c}$ is proportional to the distance between $\mathbf{x_0}$ (corresponding to $\mathbf{x_t}$) and $\mathbf{c}$, especially in the final sampling stages.

The time-dependent energy conditioning is approximated as:
\begin{equation}
     \mathcal {E}(\mathbf {c}, \mathbf {x}_t)\approx \mathcal {D}_{\boldsymbol {\theta }}(\mathbf {c}, \mathbf {x}_{0|t}). \label {eq:energy_approx} 
\end{equation}

Combining Eq. \ref{eq:energy_samp} and Eq. \ref{eq:energy_approx}, the sampling equation is:
\begin{equation}
     \mathbf {x}_{t-1} = \mathbf {r}_t - \alpha _t\nabla _{\mathbf {x}_t}\mathcal {D}_{\boldsymbol {\theta }}(\mathbf {c}, \mathbf {x}_{0|t}(\mathbf {x}_t)), \label {eq:energy_guided_sampling} 
\end{equation}

\vspace{-4mm}
\subsection{Inference-time Conditional Generation} \label{sec:color_cond}

Our first contribution is generating images conditioned on a text prompt $\mathbf{T}$ and a reference control $\mathbf{R}$ via inference-time optimizations. We interpret the diffusion model as combining the unconditional score estimator and conditional energy function (Section \ref{sec:cond_score}).

The sampling process in the latent diffusion model, including the conditional energy function, remains the same. This flexibility allows us to represent noise predictions $x_t$ from the diffusion model, incorporating the correction term as learned gradients of the Energy function (Eq. \ref{eq:energy_guided_sampling}).

Figure \ref{fig:architecture} provides an overview of our architecture. During inference, given a text prompt $T$ and a condition $R$, we sample the latent diffusion model over multiple timesteps, except during a specific interval called the \textbf{\textit{Conditioning Zone (CZ)}} detailed in Section \ref{sec:iterative_strategy}. For timesteps within $CZ \in (q,r)$, we decode the latent code to retrieve an RGB image, extract features from the denoised image and reference control, compute the loss $L$ between them, and use this to calculate the gradient of the loss. This process is repeated for $(r - q)$ timesteps, iterated $y$ times per timestep, with $y$ based on our iterative sampling strategy (Section \ref{sec:iterative_strategy}).

We conducted experiments with color and edge control, as discussed in the following subsections.


\textbf{Color. \label{sec:color}} Our method provides the ability to generate images conditioned on the text prompt and an input color palette. This is by far the first work that takes in a color palette as input and does color conditioning at test-time.


Given an input text prompt $t$ and an input color palette $p$, we first generate a spatial color palette $sp$. In the generation of the spatial color palette, a uniform distribution was employed to ensure an unbiased and evenly distributed selection of colors. The color values were randomly sampled from a continuous uniform distribution defined over the input color palette space. The lower and upper bounds of the distribution were determined based on the desired color range for the spatial representation. This approach guarantees that the resulting color palette spans the specified range in a fair and systematic manner, contributing to the diversity and even distribution of colors across the spatial visualization. 

To generate the image conditioned on the color palette, we follow the approach mentioned in Section \ref{sec:color_cond} where we compute the loss term between the spatial color palette $sp$ and the denoised image $di_t$ at each step in the Conditioning Zone. We then compute the gradient of this loss and repeat the process for $x$ timesteps which is further repeated $y$ times for each timestep.

We employ a pre-trained color encoder network to compute LAB color features from the spatial palette image $sp$ and the denoised image $di_t$. We then compute the L2 Euclidean distance measurement $\mathbf{L_{Euclidean}}$ between the two color features and use it as a loss term. This loss term helps in optimizing the color features of the denoised image to closely match those of the spatial palette image.
\begin{equation}
\text{L}_{\text{Euclidean}} = \| \text{LAB}_{gen} - \text{LAB}_{\text{ref}} \|_2 \label{eq:euclidean}
\end{equation}
where $\mathbf{\text{LAB}_{\text{gen}}}$ represents the LAB features of the generated image, and $\mathbf{\text{LAB}_{\text{ref}}}$ represents the LAB features of the reference image.

\begin{figure*}[!ht]
 \centering
 \captionsetup{type=figure}
 \includegraphics[width=\linewidth]
 {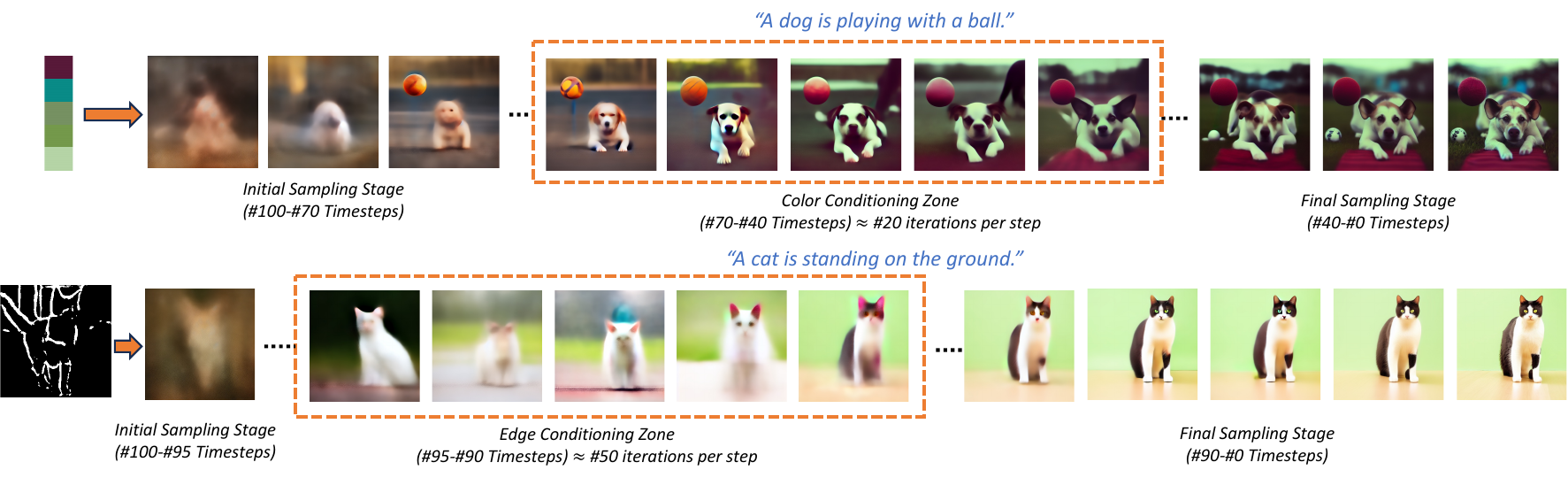}
 \caption{Demonstration of the amplified effect of applying conditional control in specific time interval for color and edge conditioning. It can be observed that the color conditioning happens in the middle stage of the sampling process whereas the edge conditioning happens in the early stage of sampling. We can observe that in the second row, the position and structure of the cat changes rapidly to fit the reference edge map. }
 \label{fig:iterative_sam}
 \vspace{-5mm}
 \end{figure*}

We introduce another innovative loss function inspired from \cite{vavilala2023applying} aimed at promoting coherence between the synthesized image's palette and the specified input palette. For a given sampling stage, let $m \in \mathbb{R}^P$ represent a probability distribution over $P$ colors in the input color palette. The colors associated with the palette are organized in a $P \times 3$ matrix $Q \in \mathbb{R}^{(P \times 3)}$. During the sampling process, we decode the noisy image $x_t$ to obtain an RGB image and then construct a pairwise distance matrix between the pixels in the denoised image and the input palette $Q$. Denoting the \(512^2 \times P\) distance matrix as \(D\) (reflecting our use of \(512^2\) dimensionality images) and employing Euclidean distance between each pixel and color in the palette, we compute the softmax along the \(P\)-dimensional palette dimension,

\vspace{-4mm}
\begin{equation}
D_{\text{sm}} = \text{softmax}(-\rho D). 
\end{equation}
Here, the sharpness parameter \(\rho\) is set to 100, encouraging a stronger affinity for the nearest palette color for each pixel. Finally, we sum \(D_{\text{sm}}\) along the pixel dimension, normalize it to obtain \(\hat{d}\), and define our color matching loss as a measure of the distribution similarity $\mathbf{L_{\text{DS}}}$, which is the cross-entropy between the predicted and the ground truth color distribution.
\begin{equation}
L_{\text{DS}} =  \text{CE}({d_{p}}, d_{x_t}) 
\end{equation}

where $\mathbf{d_{x_t}}$ and $\mathbf{d_{p}}$ are the color distribution for the denoised image and the input palette respectively.
The objective of this loss function is to optimize the alignment between the generated color distribution and the target palette distribution, contributing to the fidelity of the synthesized images.
The final loss term is as follows,

\begin{equation}
L_{\text{color}} = \lambda_1 \cdot L_{\text{DS}} + \lambda_2 \cdot L_{\text{Euclidean}}
\label{eq:final_loss}
\end{equation}
where \(\lambda_1\) and \(\lambda_2\) are the weighting coefficients for the respective losses whose values are experimentally assigned.

\textbf{Edge.} Another feature of TINTIN is the ability to generate images conditioned on a text prompt and edge maps. Given an input text prompt $\mathbf{tp}$ and a reference image $\mathbf{rf}$, goal is to generate an image $\mathbf{I}$ that follows the structure of $\mathbf{rf}$ with the content defined by $\mathbf{tp}$. We follow a similar approach as defined in Section \ref{sec:color_cond} with a changed setting. We utilize an off-the-shelf edge map generator \cite{xiang2021adversarial} to generate the edge map $\mathbf{e}$ from the reference image.

As defined in Section \ref{sec:color_cond}, we normally sample the latent diffusion model with an exception of a defined interval(see Section \ref{sec:iterative_strategy}). During this sampling interval, we decode the noisy image $\mathbf{x_t}$ to obtain an RGB image and then pass it through the edge map generator\cite{xiang2021adversarial} to get the corresponding edge map. The dimension of the obtained edge map is $1 \times 512 \times 512$ with values ranging from $0$ to $1$. To enhance the salient features, a thresholding operation is applied. Specifically, we set a threshold of $0.9$, such that all values below this threshold are set to $0$, while values equal to or exceeding $0.9$ remain unchanged. This thresholding process effectively accentuates edges and highlights regions of high intensity in the edge map. The high value of the thresholding operation is attributed to large amount of noise in the initial stages of the sampling process which is crucial in defining the structure of the image (see Section \ref{sec:iterative_strategy}).

We compute the loss based on the obtained edge maps for the reference image and the denoised image and then calculate the gradient of the loss and further repeat the sampling process. We employ the \textit{Intersection over Union (IoU)} loss as a measure to guide the generation process with respect to edge conditioning. The IoU is a metric commonly utilized in computer vision tasks to quantify the overlap between predicted and ground-truth masks. In our context, it serves as a proxy for edge alignment.

The IoU is calculated as the ratio of the intersection of the predicted and ground truth edges to their union. Specifically, for a given pixel, let $\mathbf{e_{x_t}}$ represent the predicted edge map and $\mathbf{e_{rf}}$ denote the ground-truth edge map. The IoU loss is defined as:
\begin{equation}
\text{L}_{\text{IoU}} = \frac{\text{A}(\mathbf{e_{x_t}} \cap \mathbf{e_{rf}})}{\text{A}(\mathbf{e_{x_t}} \cup \mathbf{e_{rf}})}
\end{equation}
where \(\text{A}(\cdot)\) represents the number of pixels in the corresponding set. This loss encourages the model to generate images where the predicted edges closely align with the ground truth edges, fostering improved edge fidelity during the image generation process.





\textbf{Iterative Sampling Strategy. \label{sec:iterative_strategy}} The current method struggles with precise generation under fine-grained conditions like color palettes or edge maps, especially with 100 DDIM sampling steps. This is due to weak guidance and excessive freedom in unconditional scoring, resulting in significant deviation from intended conditional control. To tackle this, we adopt an iterative sampling strategy \cite{lugmayr2022repaint, wang2023zero}, effectively tightening conditioning on the unconditional score during image generation. By iteratively refining generated images, this approach prevents deviation from intended conditional control.


The iterative sampling strategy involves revisiting the current intermediate result, $\mathbf{x_t}$, and navigating it back by $\mathbf{q}$ steps to $\mathbf{x_{t+q}}$, then resampling it back to $\mathbf{x_t}$. This method introduces additional sampling steps, refining conditioning in the generative process. By iteratively refining intermediate results, this strategy enhances alignment with desired fine-grained conditions, improving image quality and adherence to specified controls.


Examining effect at each timestep with varied repetitions, we investigate conditioning during image generation, focusing on color and edge conditions. Our analysis, informed by previous works \cite{zhang2023prospect,yu2023freedom}, reveals nuanced conditioning effects within specific time intervals, termed as the \textbf{\textit{Conditioning Zone(CZ)}} and a constant number of iterations. Determined through rigorous experimentation on the COCO dataset \cite{lin2015microsoft}, CZ marks the most effective conditioning period, as depicted in Fig.~\ref{fig:iterative_sam}.

It can be clearly observed from Figure \ref{fig:iterative_sam} that in the context of color control, our experimentation reveals optimal conditioning outcomes when applying the iterative sampling strategy between $70$ and $40$ timesteps, with a repetition number of $\approx20$. These specific values are determined after rigorous experimentation and analysis. Notably, the rationale behind applying color control in the middle stages is grounded in the observation that, by this point, a substantial portion of the structural information has already been developed. Consequently, these stages are more receptive to semantic alterations, such as changes in color, style, etc.

In the context of edge conditioning, our application of the iterative sampling strategy is strategically positioned between $95$ and $90$ timesteps, accompanied by a repetition number of approximately $\approx50$. If we observe Figure \ref{fig:iterative_sam}, the position and the structure of the cat changes drastically in the mentioned timesteps inorder to fit the reference edge map. The rationale behind initiating edge control in the early stages of the sampling process is rooted in the understanding that these initial timesteps primarily contribute in generating the foundational structure or layout information within an image. Through multiple experiments, we arrived at the determination of the optimal repetition number to achieve effective edge conditioning.


\section{Results}

\begin{table}
\vspace{-3mm}
  \centering
  \resizebox{\linewidth}{!}{%
  \begin{tabular}{lcccc}
    \toprule
    & T2I-Adapter & Control-Net & \textcolor{red}{FreeDoM} & \textcolor{red}{Ours} \\
    \midrule
    FID$\downarrow$  & $49.16$ & $24.51$ & $\textcolor{red}{67.22}$ & $\underline{\textbf{\textcolor{red}{23.91}}}$ \\
    CLIP Score$\uparrow$ & $0.28$ & $0.23$ & $\textcolor{red}{0.12}$ &  $\underline{\textbf{\textcolor{red}{0.28}}}$ \\ 
    CDS ($\times 1e-2$) $\uparrow$ & $5.55$ & $3.23$ & $\textcolor{red}{2.12}$ & $\underline{\textbf{\textcolor{red}{8.43}}}$\\
    \bottomrule
  \end{tabular}%
  }
   \vspace{-3mm}
   \caption{Quantitative comparison on COCO validation set for color control. The best results are \underline{\textbf{highlighted}}. Columns in black represent the training-dependent methods while columns in \textcolor{red}{red} are the training-free methods.}
  \label{tab:color_quant}
\end{table}


\begin{table}
  \centering
  \resizebox{\linewidth}{!}{%
  \begin{tabular}{lcccccccc}
    \toprule
    & T2I-Adapter & Control-Net & PITI & SD & MaGIC  & \textcolor{red}{MasaCtrl} & \textcolor{red}{FreeDoM}  & \textcolor{red}{Ours} \\
    & (text+edge) & (text+edge) & (text+edge) & (text) & (text+edge) & \textcolor{red}{(text+edge)} & \textcolor{red}{(text+edge)} & \textcolor{red}{(text+edge)} \\
    \midrule
    FID$\downarrow$  & $19.73$ & $20.65$ & $21.20$ & $24.68$ & $22.35$ & $\textcolor{red}{28.71}$ & $\textcolor{red}{26.32}$   & $\underline{\textbf{\textcolor{red}{18.34}}}$ \\
    CLIP Score$\uparrow$ & $0.26$  & $0.25$ & $0.21$  & $0.26$ & $0.23$ & $\textcolor{red}{0.19}$  & $\textcolor{red}{0.21}$  & $\underline{\textbf{\textcolor{red}{0.26}}}$\\ 
    SSIM$\downarrow$  & $0.41$ & $0.43$ & $-$ & $-$ &  $0.45$ & $\textcolor{red}{0.21}$ & $\textcolor{red}{0.27}$  &  $\underline{\textbf{\textcolor{red}{0.40}}}$ \\
    MSE$\uparrow$ & $\underline{\textbf{0.21}}$ & $0.18$ & $-$ & $-$ &  $0.21$ &  $\textcolor{red}{0.23}$ & $\textcolor{red}{0.25}$  & $\textcolor{red}{0.19}$\\ 
    
    \bottomrule
  \end{tabular}%
  }
  \caption{Quantitative comparison on COCO validation set for edge control. The best results are \underline{\textbf{highlighted}}.Columns in black represent the training-dependent methods while columns in \textcolor{red}{red} are the training-free methods.}
  \label{tab:edge_quant}
  \vspace{-4mm}
\end{table}

Here, we compare our training-free method TINTIN with training-dependent methods like T2I-Adapter \cite{mou2023t2iadapter}, Control-Net \cite{liu2023control}, PITI \cite{wang2022pretraining}, and MaGIC \cite{wang2024magic} as well as training-free approaches like FreeDoM \cite{yu2023freedom} and MasaCtrl \cite{cao2023masactrl} for both color and edge guidance. We evaluate these methods on the COCO validation set comprising $5000$ images, using the provided captions as text prompts. For color control, we extract the top $5$ dominant colors from ground-truth images using Color-Thief. For edge control, we generate edge maps from ground-truth images using an off-the-shelf edge map generator\cite{xiang2021adversarial}.

\begin{figure*}[t]
     \centering
         \includegraphics[width=0.9\linewidth,height=9cm]{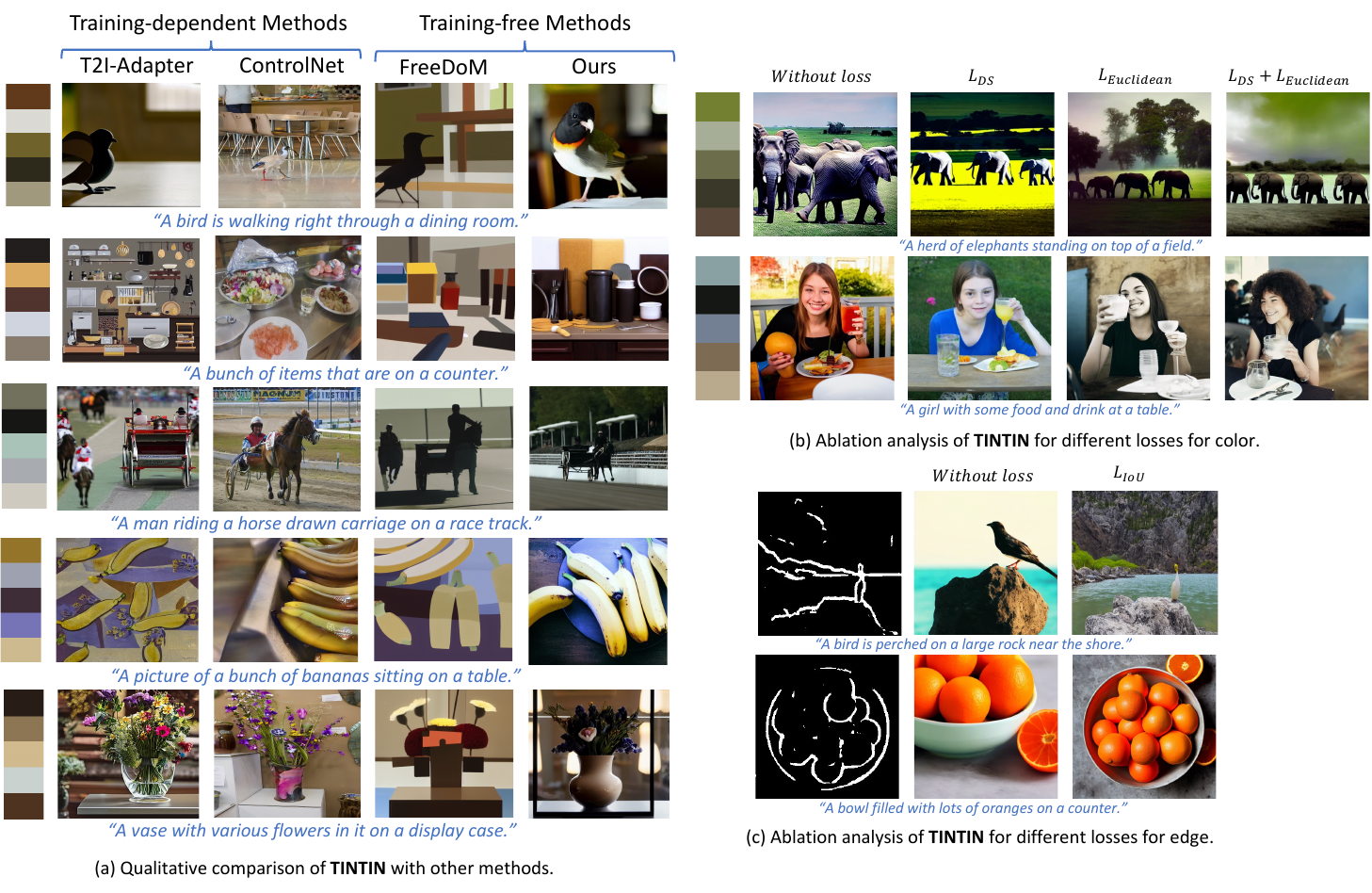}
        \caption{(a) We illustrate the ability of TINTIN in generating color palette conditioned images against trainable methods like T2I-Adapter and ControlNet and training-free methods like FreeDoM. (b) Ablation analysis of our method (TINTIN) for color conditioning. (c) Ablation analysis of our method (TINTIN) for edge conditioning.}
        \label{fig:qual_res_color}
        \vspace{-4mm}
\end{figure*}

\begin{figure}[!ht]
     \centering
         \includegraphics[width=\linewidth,height=6cm]{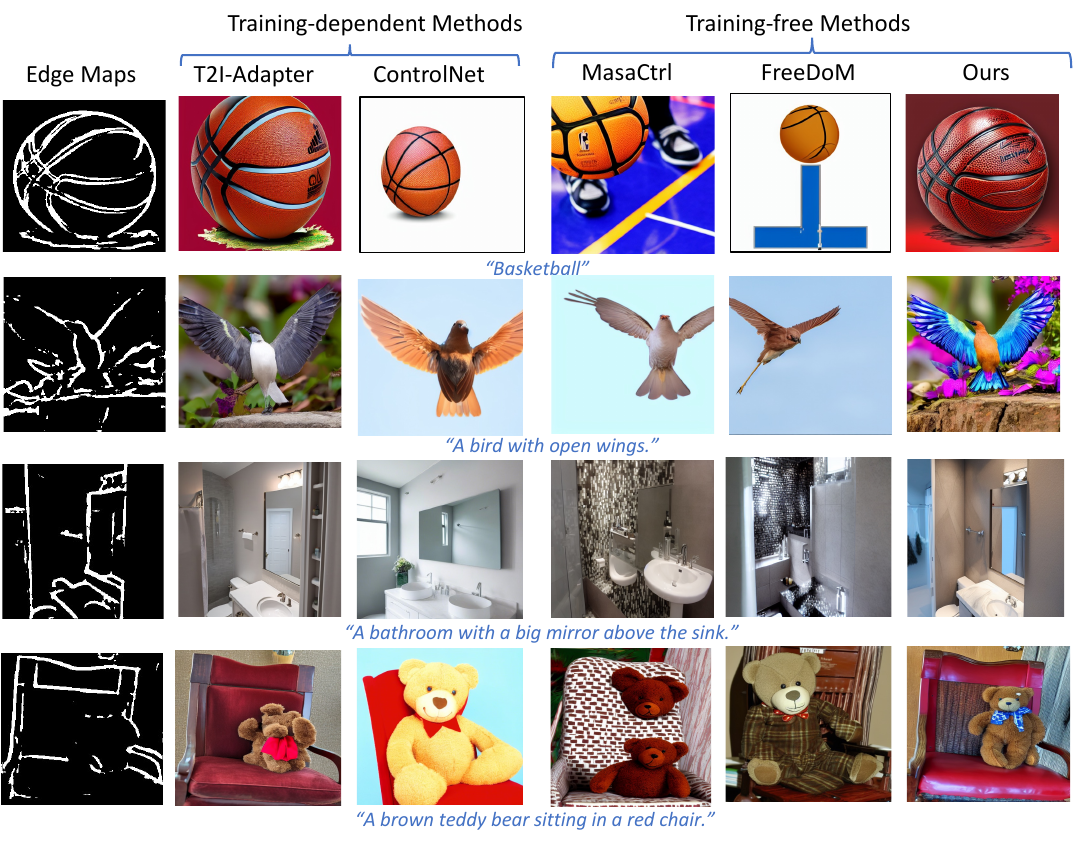}
        \caption{We compare TINTIN's edge map conditioned image generation against trainable methods (T2I-Adapter and ControlNet) and training-free methods (MasaCtrl and FreeDoM). TINTIN exhibits superior diversity in image generation, closely following the structure of the reference edge map, outperforming other methods.}
        \label{fig:qual_res_edge}
        \vspace{-7mm}
\end{figure}



As seen in Figure \ref{fig:qual_res_color}(a), TINTIN produces images with superior color distribution from the palette compared to other methods. Figure \ref{fig:qual_res_edge} illustrates that TINTIN generates high-quality images that adhere closely to the reference edge map structure, surpassing other training-free methods while matching the performance of training-dependent ones. TINTIN's edge conditioning provides a structural outline from the reference, allowing flexibility for detailed generation, striking a balance between structure guidance and detail flexibility.

\textbf{Quantitative Results.} We evaluate the effectiveness of our training-free approach against various trainable and training-free methods by assessing image quality and conditional control. We generate images using $7$ random seed values for all methods and average the scores for each prompt-condition pair. Results are summarized in Tables \ref{tab:color_quant} and \ref{tab:edge_quant}.


To evaluate image generation, we compute the Fréchet Inception Distance (FID) score \cite{Seitzer2020FID} between ground truth and generated images, and the CLIP Score (ViT-L/14) \cite{radford2021learning} measuring similarity between generated images and input text prompts. Additionally, we compute the Color Distribution Score (CDS) to assess color distribution. CDS is obtained by extracting the top 5 colors from generated images using Color-Thief \cite{colorthief} and computing the Jaccard Similarity Coefficient \cite{jaccard1901etude} with the ground-truth palette. As shown in Table \ref{tab:color_quant}, TINTIN surpasses other methods across these metrics, highlighting its efficacy in achieving superior fidelity and color distribution in generated images.

\begin{table}
\vspace{-3mm}
  \centering
  \resizebox{\linewidth}{!}{%
  \begin{tabular}{l|cccc|cc}
    \toprule
    & \textbf{\textit{Without Loss}}  & $\mathbf{L_{DS}}$ & $\mathbf{L_{Euclidean}}$ & $\mathbf{L_{DS}+L_{Euclidean}}$ & \textbf{\textit{Without Loss}} & $\mathbf{L_{IoU}}$  \\
     & (text+color) & (text+color) & (text+color) & (text+color)  & (text+edge) & (text+edge) \\
    \midrule
    FID$\downarrow$ & $34.15$   & $55.34$ & $29.73$  & $\mathbf{23.91}$ & $32.11$ & $\mathbf{18.34}$ \\
    CLIP Score$\uparrow$ & $0.19$  & $0.13$ & $0.20$  & $\mathbf{0.28}$ & $0.19$ & $\mathbf{0.26}$  \\ 
    CDS ($\times 1e-2$)$\uparrow$ & $1.34$ & $4.21$ & $5.27$ &$\mathbf{8.43}$ & $-$ & $-$ \\
    SSIM$\uparrow$  &  $-$ & $-$ &  $-$ & $-$ & $0.18$ & $\mathbf{0.40}$  \\
    MSE$\downarrow$ & $-$ & $-$  &  $-$ & $-$ & $0.29$ & $\mathbf{0.19}$  \\ 
    \bottomrule
    
  \end{tabular}%
  }
\vspace{-2mm}
\caption{\textbf{Ablation analysis} for our method quantifying the effect of various components.}
  \label{tab:ablation_tab}
  
\end{table}

\begin{table}
\vspace{-2mm}
  \centering
  \resizebox{\linewidth}{!}{%
  \begin{tabular}{l|ccccc}
    \toprule
    & \textbf{T2I-Adapter}  & \textbf{Control-Net} & \textbf{FreeDoM} & \textbf{MasaCtrl} & \textbf{Ours}   \\
    \midrule
    Training Time & $~3 days$   & $~5 days$ & $0.0$  & $0.0$ & $\textbf{0.0}$  \\
    Average Inference Time & $\textbf{~8 secs}$  & $~17 secs$ & $~18 secs$  & $~22 secs$ & $\textbf{~15 secs}$  \\ 
    Average Total Time (Approx) & $3 days+8 secs$ & $5 days+17 secs$ & $18 secs$ & $22 secs$ & $\textbf{15 secs}$  \\
    FLOPs & $0.88 T$ & $0.99 T$ & $0.68 T$ & $-$ & $\textbf{0.63 T}$  \\
    \bottomrule
  \end{tabular}%
  }
  \caption{\textbf{Runtime comparison} between our method and other baseline methods. Training-dependent methods were trained on 164K images from the COCO dataset using 4 NVIDIA Tesla 40G-A100 GPUs. Inference time represents the average duration for 100 sampling steps over 5K images. Here, \textit{secs} denotes seconds. FLOPs stands for Floating Point Operations per Second.}
  \label{tab:timing}
  \vspace{-5mm}
\end{table}


In edge control, we compute the Structural Similarity Index (SSIM) and Mean Squared Error (MSE) between the ground-truth image's edge map and the generated image. As indicated in Table \ref{tab:edge_quant}, our approach consistently outperforms others, except for MSE, where it achieves comparable results to T2I-Adapter \cite{mou2023t2iadapter}, demonstrating the effectiveness of our training-free methodology.


\textbf{Ablation Study.} We conduct a comprehensive ablation analysis to evaluate parameter efficacy. For color conditioning, we assess results generated without losses, with only distribution similarity ($\mathbf{L_{\text{DS}}}$) or only Euclidean ($\mathbf{L_{Euclidean}}$) loss, and with both losses combined (see Section \ref{sec:color}). As shown in Table \ref{tab:ablation_tab} and Figure \ref{fig:qual_res_color}(b), the combined losses notably outperform others in terms of image quality, coherence, and color distribution fidelity (further details in Supplementary). Similarly, for edge conditioning, we compare results without loss and with Intersection over Union (IoU) loss ($\mathbf{L_{\text{IoU}}}$). Findings, summarized in Table \ref{tab:ablation_tab} and illustrated in Figure \ref{fig:qual_res_color}(c), show a significant improvement in edge conditioning and overall generation quality with the incorporation of the loss function. We have included additional results in the supplementary material, including an ablation study that examines the impact of selecting different timesteps for various controls.

\textbf{Runtime Analysis.} We compare the runtime of our training-free approach with other baseline methods (see Table \ref{tab:timing}). Our method exhibits significantly faster performance than training-dependent methods, while also surpassing training-free approaches with a large time difference.

\textbf{User Study.} To evaluate conditional methods' efficacy on COCO dataset samples, we conducted a user study with $65$ participants. Each participant assessed $8$ generated samples from T2I-Adapter, ControlNet, and our method, randomly selected from the COCO validation set. Participants compared color distribution to provided palettes and selected the closest match. Results showed a preference for images generated by our method, with $68.43\%$ of participants favoring our approach. This highlights our method's effectiveness in achieving desirable color distributions.


\section{Conclusion}

We introduce TINTIN, a novel training-free approach for conditional image generation using text-to-image diffusion models. TINTIN leverages energy function gradients to guide generation towards provided conditions during sampling. Notably, it pioneers color palette control in test-time text-to-image synthesis. While showcasing its efficacy in color and edge control, TINTIN's applicability extends to diverse conditioning types such as style, segmentation maps, and pose. We propose various loss functions for conditioning and conduct comprehensive quantitative and qualitative analyses on the COCO validation dataset. Even though our method is faster than the existing state-of-the-art approaches, we further wish to optimize the time required during inference as part of future work. 

%
\section{Appendix}
In Section \ref{sec:results}, we show additional qualitative results for our method against some state-of-the-art training-required methods like T2I-Adapter \cite{mou2023t2iadapter} and Control-Net \cite{liu2023control}. In section \ref{sec:details}, we provide some additional implementation details about our method.

\begin{figure}[!ht]
     \centering
         \includegraphics[width=\linewidth]{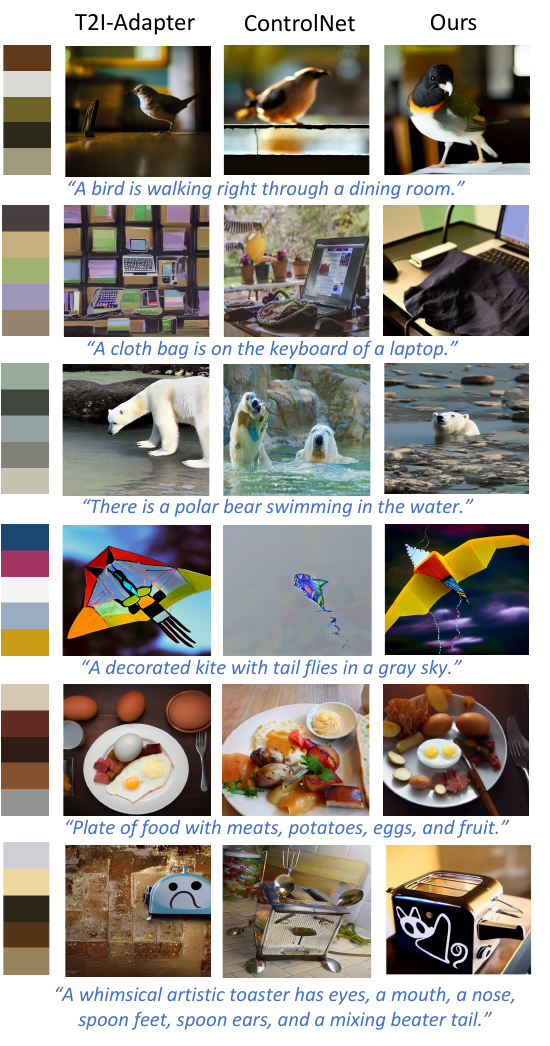}
        \caption{We illustrate the ability of TINTIN in generating color palette conditioned images against trainable methods like T2I-Adapter\cite{mou2023t2iadapter} and ControlNet\cite{liu2023control}. Our training-free approach is able to generate color balanced results as compared with other state-of-the-art methods that require training a model.}
        \label{fig:supple_color}
\end{figure}

\begin{figure}[!ht]
     \centering
         \includegraphics[width=\linewidth]{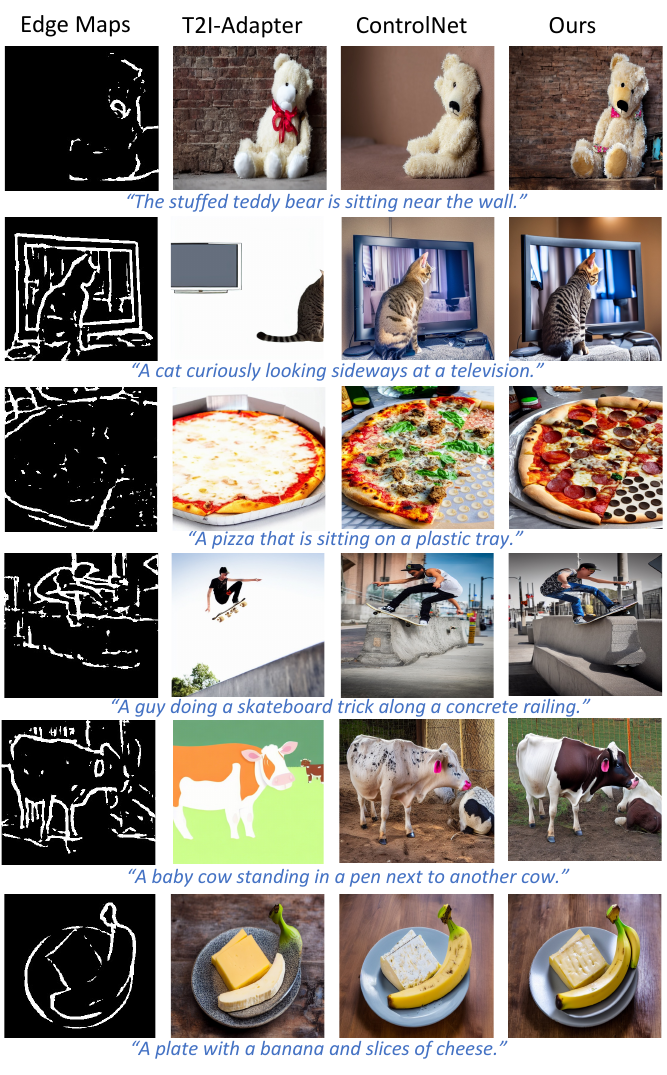}
        \caption{We illustrate the ability of TINTIN in generating edge map conditioned images against trainable methods like T2I-Adapter\cite{mou2023t2iadapter} and ControlNet\cite{liu2023control}. Our training-free approach is able to generate diverse images following the structure of the reference edge map as compared with other state-of-the-art methods that require training a model.}
        \label{fig:supple_edge}
\end{figure}


\subsection{Qualitative Results} \label{sec:results}

As depicted in Figure \ref{fig:supple_color}, TINTIN exhibits a superior capability in producing images that closely align with the color distribution inherent in the provided color palettes, surpassing alternative methodologies. Notably, the generated images by TINTIN demonstrate a heightened fidelity to the given prompt in comparison to the outputs from T2I-Adapter \cite{mou2023t2iadapter} and Control-Net \cite{liu2023control}. In Figure \ref{fig:supple_edge}, we present a visual comparison of images generated by our approach alongside those generated by the aforementioned methods, each conditioned on corresponding edge maps. Evidently, TINTIN-generated images exhibit a commendable adherence to the structural cues provided by the edge maps, outperforming alternative approaches in this regard. 
We observe that our method not only performs better than some training-dependent approaches but is also superior than many training-free methods.

\subsection{Implementation Details} \label{sec:details}
In our approach, we leverage the pre-trained Stable Diffusion model \cite{rombach2022highresolution}, specifically version 1.4, which is also employed in the baseline methods. We use a A100-8GB GPU machine for our experimentations. During the sampling process, we uniformly resize both the denoised image and the condition maps to dimensions of $512 \times 512$. In the context of color conditioning, we determine the optimal values for \(\lambda_1\) and \(\lambda_2\), utilized in the final loss term as outlined in Equation 11 of the main paper, to be $1$ and $0.1$ respectively. The reported results in our study are grounded in the application of these specified parameter values.  

\subsection{Limitations}
\label{subsec:limitations}

In this section, we briefly discuss a few limitations of TINTIN when seen in a conditional text-to-image generation setup. Firstly, TINTIN is slower than training-required methods during inference due to the presence of gradient computation of the energy function and iterative sampling strategy. Secondly, since the performance of TINTIN is highly dependent on the loss functions and the designated Conditioning Zone (CZ) for a specific condition, the extension of TINTIN to other conditions become dependent on loss functions and the conditioning zone that can be very different for different conditions. Moreover, since we are building on top of existing text-to-image models, any potential fairness considerations for these base models will flow to our method as well.

\bibliography{aaai25}

\end{document}